\documentclass[conference]{IEEEtran}
\IEEEoverridecommandlockouts
\usepackage{amsmath,amssymb,amsfonts}
\usepackage{graphicx}
\usepackage{textcomp}
\usepackage{xcolor}
\usepackage[utf8]{inputenc}

\usepackage{algorithmicx}
\usepackage[ruled]{algorithm}
\usepackage[noend]{algpseudocode}
\usepackage{amsmath}
\usepackage{lipsum}
\usepackage{graphicx}
\usepackage{caption}
\usepackage{subcaption}
\usepackage{longtable}
\graphicspath{{figures/}}
\usepackage{titlesec}

\usepackage{subcaption}
\usepackage{schemabloc,tikz}
\usepackage{amsfonts,epsfig,array,multirow,graphicx,amsmath,amsthm,ltablex,tabularx,setspace,arydshln,amssymb,multirow}
\usetikzlibrary{circuits, arrows}

\usepackage[utf8]{inputenc}
\usepackage{amsmath}
\usepackage{amsfonts}
\usepackage{amssymb}
\usepackage{bbm}
\usepackage{mathtools}
\usepackage{textcomp}
\usepackage{stackengine}
\usepackage{booktabs}
\usepackage{longtable}
\usepackage{multirow}
\usepackage{graphicx}
\usepackage{subcaption}

\patchcmd{\subequations}{\def\theequation{\theparentequation\alph{equation}}}%
{\def\theequation{\theparentequation.\arabic{equation}}}{}{}

\usepackage{biblatex}
\def\BibTeX{{\rm B\kern-.05em{\sc i\kern-.025em b}\kern-.08em
    T\kern-.1667em\lower.7ex\hbox{E}\kern-.125emX}}
\usepackage{hyperref}
\hypersetup{
    colorlinks=true,
    linkcolor=blue,
    filecolor=magenta,      
    urlcolor=cyan,
    citecolor=blue
}
\makeatletter
\newcommand{\algmargin}{\the\ALG@thistlm}

\makeatother

\algnewcommand{\parState}[1]{\State%
    \parbox[t]{\dimexpr\linewidth-\algmargin}{\strut\hangindent=\algorithmicindent \hangafter=1 #1\strut}}
    
\begin{document}

\title{Local NMPC on Global Optimised Path for Autonomous Racing}

\author{\IEEEauthorblockN{Dvij Kalaria \textsuperscript{*1}, Parv Maheshwari \textsuperscript{*1}, Animesh Jha \textsuperscript{*1}, Arnesh Kumar Issar \textsuperscript{*1},\\ Debashish Chakravarty \textsuperscript{1}, Sohel Anwar \textsuperscript{2}, Andres Tovar \textsuperscript{2} \thanks{$^*$ Equally contributing authors} \thanks{$^{1}$ Indian Institute of Technology, Kharagpur \{dvij, parvmaheshwari2002, jha.animesh01, arnesh.issar\}$@$iitkgp.ac.in, dc@mining.iitkgp.ac.in } \thanks{$^{2}$ IUPUI \{soanwar,tovara\}$@$iupui.edu}}
}

\maketitle

\begin{abstract} 
The paper presents a strategy for the control of an autonomous racing car on a pre-mapped track. Using a dynamic model of the vehicle, the optimal racing line is computed, taking track boundaries into account. With the optimal racing line as a reference, a local nonlinear model predictive controller (NMPC) is proposed, which takes into account multiple local objectives like making more progress along the race line, avoiding collision with opponent vehicles, and use of drafting to achieve more progress.
\end{abstract}
\section{Introduction}
Autonomous driving is an active field of research that poses challenges in various fields, including Motion Planning, Controls, and Perception. The task can be as simple as lane following and goal-reaching or complex as multi-agent driving scenarios. Finding optimal trajectories that are feasible is a key challenge associated with this field. However,  for low-speed autonomous driving, these challenges can be solved relatively easily, and there is extensive work on the same. On the other hand, high-speed autonomous driving, although discussed in many works, is not heavily researched due to the challenging experimental setups and safety aspects of high-speed driving. With the advent of various competitions like Roborace and the ongoing Indy Autonomous Challenge (IAC), the field of high-speed autonomous racing is bound to attract a lot more attention. Various small-scale setups are also established for experimental work in this field which can help standardize this field for future research. 
This paper describes the design and implementation details of an optimization based planning and control algorithm used for autonomous racing in a pre-mapped environment as a 2 stage process. We first approximate the global optimal trajectory and then use it for local Nonlinear Model Predictive Control (NMPC) with multiple objectives, including inter vehicle collision avoidance. This work was carried out as a part of the Indy Autonomous Challenge (IAC). All algorithms described in this paper were simulated and tested using ANSYS VRXPERIENCE simulator with the vehicle models and scenarios provided by the IAC team. This paper does not concern itself with localization and perception in high speed scenarios and assumes the availability of the accurate position of the ego vehicle and the opponent vehicles within a certain range admissible for the RADAR or LIDAR sensors to detect. The paper makes the following novel contributions
\begin{enumerate}
    \item Use of optimal racing line as a reference for NMPC as opposed to the conventional use of center line
    \item Dynamically modelling the tire forces at high speeds using a modified Pacejka model.
    \item Incorporating a drafting effect model within the control scheme for overtaking scenarios.
    \item The proposal to merge with the reference line at turns which gets more progress along the track
\end{enumerate}

\section{Related works}\label{Previous works}

Modelling vehicle dynamics using kinematic bicycle models for Model Predictive Control was studied by \cite{Kinematic_and_dynamic_vehicle_models}, a dynamic bicycle model with nonlinear tire forces was studied by \cite{kabzan2019amz}. Tire forces are generally modelled by the Pacejka tire model \cite{Pacejka1991THEMF}, however most previous works \cite{kabzan2019amz} use a simplified model and do not consider the dynamic nature of the model at higher speeds. The presence of a pre-mapped environment allows us to use an optimised reference line instead of simply using the centerline. Obtaining a minimum time race-line has been studied by \cite{doi:10.1080/00423114.2019.1704804} where they solve a global optimization problem to obtain the race-line.

The algorithms used for local motion planning in the context of autonomous racing can be divided into three classes: 1) Sampling-based approaches 2) Graph-based approaches 3) Numerical optimization based approaches. Sampling based/incremental search approaches finds their strength under unstructured or unknown environment \cite{Karaman2010OptimalKM}. Graph-based approaches span a spatial or spatio-temporal hierarchical tree along the drivable area. Commonly each node in the graph holds a cost calculated based on a heuristic. Using graph-search algorithms, the optimal path is chosen. These approaches discretize the action or spatial space in some way and choose the optimal track after exhaustively exploring all options by building a tree. They have proven to be effective in autonomous driving scenarios \cite{Stahl2019MultilayerGT} \cite{Li2017DevelopmentOA} \cite{Hu2018DynamicPP}, but however one of the key issues that they face in the case of racing is the precise and sensitive application of controls to achieve which, the level of discretization has to be increased leading to high computation time. The third class is the optimization based methods which are mostly dedicated to Model Predictive Control (MPC) \cite{Gao2010PredictiveCO} \cite{kabzan2019amz}. They  primarily focus on optimizing an objective to get the control commands. They have achieved good results in the context of autonomous racing, but is an active research topic both in terms of computation time and achieving more optimal results. Optimization based approaches for control of autonomous racing vehicles like NMPC and MPCC have been explored by \cite{Liniger_2014}, which further incorporate obstacle avoidance with the controller. However, all of these methods use the center line of the race track as a reference. We believe that using the optimal racing line in the case of pre-mapped environment can be advantageous. \cite{Stahl2019MultilayerGT} uses optimal racing line as a reference although it is a graph based approach, we propose to use the same for optimization based methods.


\section{Vehicle Model}\label{Vehicle model}
Modelling a race car is a challenging task due to nonlinear behavior at high speeds. Our vehicle model has three sub-models, a dynamic model which predicts the next state given the current state and control commands, a tire model which models the forces on the tires in high speed scenarios, and a drafting model which takes into account the changes in vehicle dynamics due to drafting.

\subsection{Dynamic model}\label{Dynamic model}
Our base model has been adapted from \cite{kabzan2019amz} and acts under the similar assumptions
\begin{itemize}
    \item The vehicle drives on a flat surface
    \item Load transfer can be neglected
    \item The longitudinal drive-train forces act on the center of gravity.
\end{itemize}  
The equation of motion can be described using 
\begin{subequations}

\begin{flalign}
    \begin{split}
        \begin{bmatrix}
                \dot X \\
                \dot Y \\
                \dot \phi \\ 
                \dot v_x\\
                \dot v_y\\
                \dot r
        \end{bmatrix}	
                &= 
        \begin{bmatrix}
            v_x \cos(\phi) -v_y \sin(\phi) \\
            v_x \sin(\phi) +v_y \cos(\phi) \\
            r \\
            \frac{1}{m} (F_{R,x}-F_{F,y}\sin(\delta)+m v_y r)\\
            \frac{1}{m} (F_{R,y}+F_{F,y}\cos(\delta)-m v_s r)\\
            \frac{1}{I_z} (F_{F,y} l_F \cos(\delta)-F_{R,y}l_R)
        \end{bmatrix}
    \end{split}\\
    \mathbf{u} &= \begin{bmatrix} \delta & D \end{bmatrix}^T\\
    \mathbf{\theta} &= \begin{bmatrix} X & Y & \phi & v_x & v_y & r\end{bmatrix}^T 
    \\
    \begin{split}
        F_{F,y} &= f_F(\alpha_F,F_{F,z})\\
        \alpha_F &= - \arctan\left(\frac{\omega l_F + v_y}{v_x}\right) + \delta, \\
        \text{and}\ F_{F,z} &= F_{s,F,z} + \frac{1}{2} C_{lF}.\rho.S.v_x^2
    \end{split}\\ \nonumber\\
    \begin{split}
        F_{R,y} &= f_R(\alpha_R,F_{R,z})\quad \\
        \alpha_R &=  \arctan\left(\frac{\omega l_R- v_y}{v_x}\right),\\
        \text{and}\ F_{R,z} &= F_{s,R,z} + \frac{1}{2} C_{lR}.\rho.S.v_x^2
    \end{split}\\ \nonumber\\
    \begin{split}
        F_{R,x} &= (C_{m1}- C_{m2}v_x)D - C_R - F_d\\
        F_d&=\frac{1}{2}v_x^2 \cdot \rho \cdot C_d \cdot S \cdot \alpha_{cd}
        \label{drag force}
    \end{split}
\end{flalign}
\end{subequations}

where the car has a mass $m$ and an inertia $I_z$, $l_R$ and $l_F$ represent the distance form the center of gravity to the rear and the front wheel respectively,\ $F_d$ is the aerodynamic drag exerted on the race car, $\alpha_{cd}$ is the coefficient introduced due to drafting (refer to Section \ref{Drafting model}), $\rho$ is air density, $S$ is the area of the race car and $C_d$ is the drag coefficient of the car.\ $F_{R,y}$ and $F_{F,y}$ are the lateral tire forces of the rear/front wheel, $F_x$ is the combined force produced by the drive-train. The state of the model $\mathbf{\theta}$, consists of $X,Y$ and $\phi$ the position and heading in a global coordinate system, as well as the $v_x, v_y$ the longitudinal and lateral velocities, and $r$ the yaw rate. The control input $ \mathbf{u}$ consists of $\delta$ the steering angle and $D$ the driving command. The driving command replicates the pedals of a driver where $D = 1$ corresponds to full throttle and $D = -1$ to full braking. $F_{s,F,z}$ and $F_{s,R,z}$ are the normal forces exerted on the front and rear tires of the vehicle when static. Further we consider the effect of the lift force exerted on the race car. Considering $C_{lF}, C_{lR}$, the coefficients of lift for front and rear, $\rho$,  air density, and $S$, the effective contact area of the race car, we can model $F_{F,z}$ and $F_{R,z}$, the vertical loads on front and rear tires at velocity $v_x$. The forces described given slip angle are modelled using functions $f_R$ and $f_F$ for front and rear tires which represent the interaction between tires and track surface calculated using the tire model in Section $\ref{Tire model}$. 

\subsection{Tire model}\label{Tire model}
The Tire-Road Interaction has been modelled using the Pacejka model \cite{Pacejka1991THEMF}. Due to the design of race-cars, tire load increases at high speeds due to a negative coefficient of lift. Race-cars work in high speed scenarios for a large part of their operation, therefore it is essential to consider  these load dependent changes. For this we have used a dynamic load dependent Magic Formula which models coefficients as a function of tire load \cite{magic_formula}. 
The steady state tire characteristic function $F_y = f(\alpha,F_z)$ calculates the lateral force $F_y$ on the tire and is based on $F_z$, the vertical load and the wheel slip angle $\alpha$ is defined in Equation \ref{eqn_for_f}. 
Using these quantities we can now model a dynamic Pacejka model as follows:
\begin{subequations}
\begin{flalign}
    df_z&=\frac{F_z-F_{z0}}{F_{z0}}\\
    S_{Hy}&=P_{Hy_1}+P_{Hy_2} \cdot df_z\\
    S_{Vy}&=F_z \cdot (P_{Vy_1}+P_{Vy_2} \cdot df_z)\\
    \alpha_y&=\alpha + S_{Hy}\\
    C_y&=P_{Cy_1}\\
    D_y&=\mu_y \cdot F_z\\
    \mu_y&=P_{Dy_1}+P_{Dy_2} \cdot df_z\\
    \begin{split}
        E_y&=(P_{Ey_1}+P_{Ey_2} \cdot df_z + P_{Ey_3} \cdot df_z^2 )\\
        &\quad \quad [1-P_{Ey_4}\cdot sgn(\alpha_y)]
    \end{split}\\
    K_{y\alpha}&=P_{Ky_1} \cdot F_{z0} \cdot \sin(2 \cdot \arctan(\frac{F_z}{P_{Ky_2} \cdot F_{z0}}))\\
    B_y&=\frac{K_{y\alpha}}{C_y D_y + \epsilon_y}
 \end{flalign}
 \begin{flalign}
    \begin{split}
        F_{y} = & D_y \sin(C_y \arctan(B_y \alpha_y - \\
        & E_y [B_y\alpha_y - \arctan(B_y\alpha_y)]))+S_{Vy} 
    \end{split}
\end{flalign}
\label{eqn_for_f}
\end{subequations}

where, $P_{Hy_1}$, $P_{Hy_2}$, $P_{Vy_1}$, $P_{Vy_2}$, $P_{Dy_1}$, $P_{Dy_2}$, $P_{Ey_1}$, $P_{Ey_2}$, $P_{Ey_3}$, $P_{Ey_4}$, $P_{Ky_1}$, $P_{Ky_2}$ are characteristic constants specific to the tire material and the road surface and are assumed to be known. $df_z$ represents the fractional change in tire load with respect to the reference tire load, $F_{z0}$, $S_{Hy}$ and $S_{Vy}$ represent offsets to the slip and lateral force in the force-slip function. $\mu_y$ is the lateral load-dependent friction coefficient. $\epsilon_y$ is a small number inserted to prevent division by zero as $F_z$ approaches zero.  

\subsection{Drafting model}\label{Drafting model}
\begin{figure}[h!]
    \centering
    \includegraphics[width=0.4\textwidth]{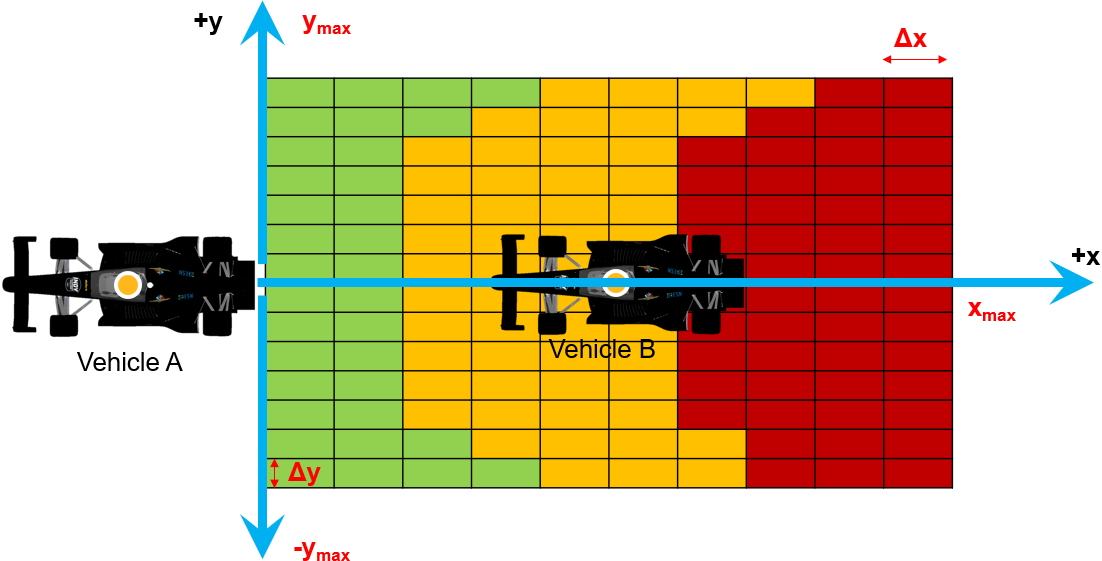}
    \caption{Area of application for drafting}
    \label{drafting}
\end{figure}

Drafting or slip streaming is an aerodynamic technique where two vehicles or other moving objects are caused to align in a close group, reducing the overall effect of drag on the preceding vehicle due to exploiting the lead object's slipstream. The effects of drafting on vehicle dynamics in the context of autonomous racing has been studied in \cite{en13195183}
We use similar formulation to express the draft forces as a reduction of the aerodynamic drag forces. In the Fig. \ref{drafting} if the preceding vehicle which is vehicle B in this case, is in the “area of application” of the vehicle A, a “draft force” is applied on vehicle B. As in Section \ref{Dynamic model}, “normal drag” slows the car and is applied in $\textbf{–}X$, the “drafting-induced drag force” should be applied in $+X$. Mathematically we express drafting by multiplying a coefficient $\alpha_{cd}$ to the normal drag force ($F_d$ in Equation \ref{drag force}) experienced by the vehicle.
\begin{flalign}
    \begin{split}
        \alpha_{cd} &= 1 +(\frac{v}{v_{max}})\cdot( \beta_{cd}-1),\\
        &\text{where}\ \beta_{cd} = \min(k_c+k_x\cdot \Delta x+k_y \cdot \Delta y,1) 
    \end{split}
\end{flalign}
\section{Methodology}

Assuming that we have a pre-mapped track with accurate lane boundaries and static obstacles, if any, we propose to divide the problem into two parts: 1) generating the optimal racing line using the known lane boundaries and well defined track and 2) using local MPC with the optimal racing line and sensor data as input for considering opponent vehicle collision avoidance and other objectives. We propose an objective function for the local MPC, which uses optimal racing line to estimate the optimal step towards lap completion and other strategies like leveraging the use of drafting and constraints such as staying within the lane limits and collision avoidance with opponents' vehicles.

\subsection{Calculating global racing line}
The first step in planning is to get the optimal global racing line. We need to get the time optimal trajectory as described in \cite{doi:10.1080/00423114.2019.1704804} following the optimization problem described in Equation \ref{global trajectory}, where $\theta_k$ is the state, $u_k$ is the action command, and $l(\theta _k,u_k)$ is the increase in progress along the center line for the time step $k$. $\theta_{start}$ is the start state which can be kept as the start condition within the admissible states $X_l$, and $F$ is the vehicle model function described in Section \ref{Vehicle model}. This ultimately maximizes the distance travelled along the center reference line and thus minimizing time. The admissible set $X_l$ encapsulates the lane boundaries and static obstacles, if any. We have used the Global Race Trajectory Optimization package provided by TUMFTM. To this package, we gave the following inputs: track information, lateral and longitudinal accelerations limit due to friction and the model parameters. The output of this package is an approximate time optimal racing line on the given race track in from which we use an array in $\mathbb{R}^{N \times 4}$              
where $N$ is the number of points depends on the stepsize and track length. The 4 columns are structured as follows: X-coordinate, Y-coordinate, target velocity $V_x$, and target velocity $V_Y$ for a point. 
\begin{equation}
    \begin{split}
    \mathop{\max}\limits_{\theta , u}&\quad\sum_{k=0}^{N-1}l_k(\theta_k,u_k) \\ 
    s.t.&\quad\theta_{0} - \theta_{start} = 0\\
    &\quad \theta_{k+1} - \Delta t F(\theta_{k},u_k) = \theta_k\\
    &\quad \theta_k \in X_l 
    \end{split}
    \label{global trajectory}
\end{equation}

\subsection{Local planner using NMPC} 
We propose to formulate a Nonlinear Model Predictive Control with the optimal trajectory, local obstacles, and opponent vehicle positions as input. Further, we consider the twin motives of making maximum progress along the track while avoiding collisions with other vehicles and at the same time leveraging the advantages of local factors like drafting (Section \ref{Drafting model}) to gain speed due to spatially local region created with reduced air resistance. However, most of the works have proposed a two level solution to this with a high-level planner that outputs a traversable trajectory and a low-level controller to track that trajectory as described under previous works (Section \ref{Previous works}). We propose a single-step solution that merges both the steps, i.e., it plans a trajectory by solving an optimization problem and taking the first planned control command as output. For solving the nonlinear problem, we use Interior Point Optimization (IPOPT). 
\subsubsection{Overtaking strategy} \label{Overtaking Strategy}
\begin{figure}[h]
    \centering
    \includegraphics[width=0.4\textwidth]{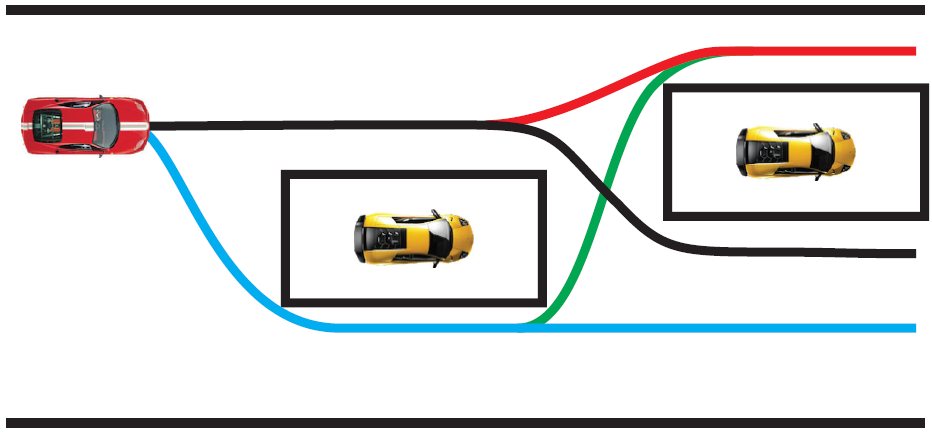}
    \caption{Combinatorial nature of the overtaking problem, as it is possible to overtake each opponent on the left or the right side \cite{Liniger_2014}}
    \label{overtaking}
\end{figure}

The primary condition for the optimality of non-linear MPC is the convexity of the loss function. However, in the case of overtaking, the optimization problem becomes non-convex as there exists multiple minima due to existence of multiple sides to overtake against the target vehicle. The local minima obtained by interior point optimization (IPOPT) depends on the side of the initial trajectory chosen. Also, with more number of opponents or local obstacles, the decision becomes combinatorial in nature. The situation is well depicted in Fig. \ref{overtaking}. One solution to this problem is to use some heuristic to determine the topology or use a Dynamic Programming solution as depicted in \cite{Liniger_2014} to get the optimal combination. However, we propose a more straightforward approximate solution by using the planned trajectory from the previous step to determine the class. If the distance with an opponent vehicle is less than a certain threshold $T_d$, we choose two combinations for the initial trajectory, one from the left of the target vehicle and the other from the right. We optimize both initial trajectories separately using the NMPC problem described later. This approach can be fully parallelized on a multi-core system. The same solution can be extended based on the amount of computation and parallel cores available to include even more classes. For getting the initial trajectories, we use the pursuit algorithm \cite{follow_the_carrot} with target positions as $X_t - d$ and $X_t + d$ where $X_t \in \mathbb{R}^2$ is the position of the target vehicle at time step $t$ and $d$ is the displacement vector with directions perpendicular to the target vehicle, and length can be chosen proportionately as per the vehicle dimensions. These initial trajectories are then used for further optimization.
\subsubsection{Cost functions}

The objectives to be considered by NMPC are: 1) making locally maximum progress along the track using the global racing line as the reference, 2) avoiding collision with opponent vehicles and other local static or dynamic obstacles, 3) using drafting effect to increase chances of overtaking the vehicle ahead, and 4) avoiding longitudinal or lateral slipping of the vehicle or any other unsafe state and the constraint to stay within the lane boundaries.

\begin{figure}[h]
    \centering
    \includegraphics[width=0.4\textwidth]{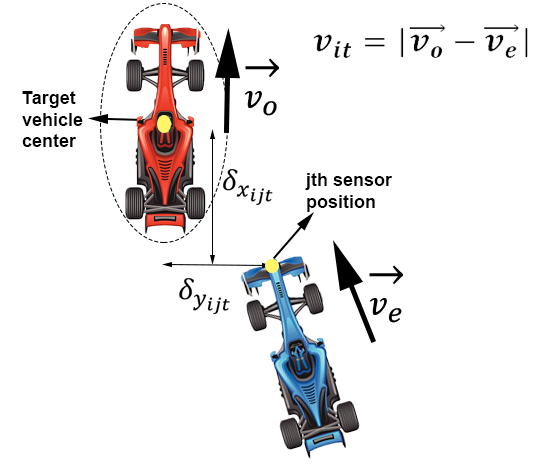}
    \caption{The reference frame for $\delta x_{i,j,t}$ \ \text{and} \ $\delta y_{i,j,t}$}
    \label{obs_avoidance}
\end{figure}

There has been a substantial study into deriving a strategy to maximize progress along the reference line using local MPC. One such work is \cite{Liniger_2014}, which proposes a trade-off between tracking the reference trajectory by minimizing the contouring error, which they define as the perpendicular distance from the reference line, and the maximization of progress, which they define as distance along the trajectory from the corresponding perpendicular projection points from the start and end position of the locally planned trajectory. However, we argue that this fits well when the reference line is the center line as they have used. However, since we are using the globally optimal trajectory as the reference, we propose a novel cost term that incorporates both the tracking of the race line and what we believe as the optimal way to merge with the race line only at the time of turns to not interfere with the overtaking or obstacle avoidance objective. First, we define the projection function $P$ as defined in Equation \ref{projection} that finds the projection point on the reference line. However, this itself is an optimization problem; hence we find the approximate projection point by starting with the projection point along the X-axis with a reference frame located at the ego vehicle's center position and optimizing with $r$ no. of iterations. Choosing $r=3$ gave empirically very close projection point. The cost term is as defined in Equation \ref{cost function}, where $v_t$ denotes the speed of the ego vehicle at time step $t$, $\theta_t$ denotes the heading of the ego vehicle at time $t$, $\alpha$ denotes the parameter for perpendicular projection point on the approximate spline of the race line, $\theta(\alpha)$ and $R(\alpha)$ are the tangential angle and the radius of curvature at the perpendicular point. Thus, this term would put more weight on merging with race-line only during turns when the radius of curvature of the projection point is less. The comparison in the results when the contouring error cost term was used and when the new cost term was used has been carried out in Section \ref{Results}.
\begin{equation}
        P(X, Y) \triangleq\mathop{\min}\limits_{\theta}(X - X^{ref}(\theta))^2 + (Y - Y^{ref}(\theta))^2 
        \label{projection}
\end{equation}
\begin{equation}
    \begin{split}
    C_{pt} &= \frac{v_t \cos(\theta_t-\theta(\alpha)) R(\alpha)}{R(\alpha) + |D_k|},\\
    & \text{where}\ D_k = \sqrt{(X-X(\alpha))^2 + (Y-Y(\alpha))^2},\\
    & \text{and} \ \alpha = P(x_t,y_t)
    \end{split}
    \label{cost function}
\end{equation}
\begin{figure}[h]
    \centering
    \includegraphics[width=0.3\textwidth]{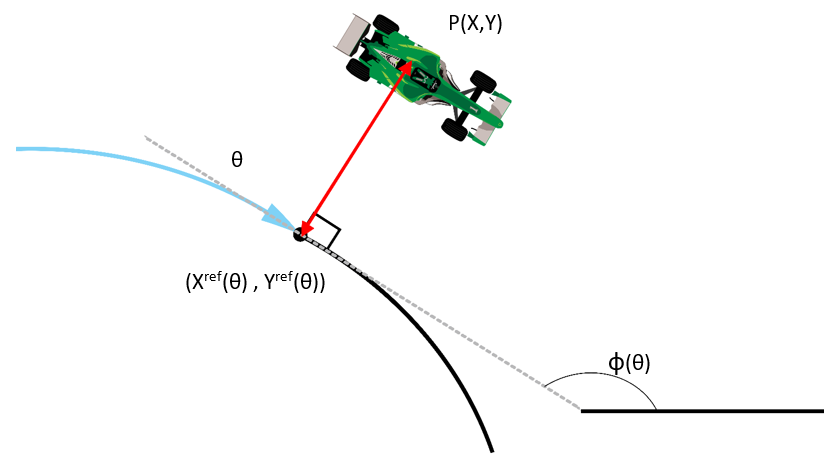}
    \caption{The projection point on the optimal racing line}
\end{figure}
The second objective is to avoid collision and maintain a safe distance for opponent vehicles and static obstacles if any. This has been studied extensively in prior works. The method in \cite{Gao2010PredictiveCO} sets up a cost term in the objective function to penalize coming close to the obstacle with a penalty inversely proportional to the distance from the obstacle and directly proportional to the speed. However, as we have dynamic obstacles, we replaced the speed of the vehicle with the relative velocity magnitude between the opponent vehicle and the ego vehicle, and instead of Euclidean distance, we use the distance scaled with different proportions along the $x$ and $y$-axis proportionate to the opponent vehicle dimensions such that we have ellipsoidal equivalent region. The resultant cost term at time step $t$, for target vehicle or obstacle $i$ and sensor $j$, is described in Equation \ref{second objective} where ($\delta x_{i,j,t}$,$\delta y_{i,j,t}$) denotes the position of the sensor $j$ with respect to the local coordinate frame of the target vehicle as shown in Fig. \ref{obs_avoidance}. $v_{i,t}$ denotes the relative velocity magnitude with the $i^{th}$ target vehicle detected at time step $t$, $l_x$ and $l_y$ denote the length and width of the target vehicle respectively. $M$ and $k$ denote the number of target vehicles and sensors, respectively. The term is also multiplied with the sigmoid term to ensure non-overlap and smooth transition from the drafting objective to the obstacle avoidance objective. $S_{do}$ and $K_{do}$ are constants that determine the smoothness of the transition and the distance at which the transition must occur.
\begin{equation}
    \begin{split}
    & C_{ot} = \sum_{i=0}^M \sum_{j=0}^k \frac{v_{i,t}}{(\frac{\delta x_{i,j,t}}{l_x})^2 + (\frac{\delta y_{i,j,t}}{l_y})^2} \sigma(S_{do}(K_{do} - \delta x_{i,j,t})),\\
    & \quad \quad \quad \quad \quad \quad \quad \quad \quad \text{where}\ \sigma (z) = \frac{e^{z}}{1 + e^{z}}
    \end{split}
    \label{second objective}
\end{equation}
The third objective is to use the drafting effect, which is the spatially local regions created behind the opponent vehicles where a reduced wind force is experienced. To consider this, we propose the cost term as described in Equation \ref{third objective} where $M$ is the number of vehicles and ($\delta x_{i,t}$,$\delta y_{i,t}$) denotes the position of the ego vehicle with respect to the local coordinate frame of the target vehicle $i$ at time step $t$. This objective term would attract the vehicle behind a certain vehicle and thus leverage the use of drafting. Also, the term is multiplied by an activation term with opposite activation to the obstacle avoidance term to smoothly transition back to the obstacle avoidance objective when the vehicle comes close to the target for overtaking. The path followed by the vehicle before and after the drafting term has been demonstrated in Section \ref{Results}.
\begin{equation}
C_{dt} = \sum_{i=0}^M|\delta y_{i,t}| (1 - \sigma(S_{do}(K_{do} - \delta x_{i,j,t})))
\label{third objective}
\end{equation}
For additional safety, we have used an extra penalty cost term adapted from \cite {kabzan2019amz}, which accounts for the lateral slip angles ($\alpha_F$ and $\alpha_R$) going beyond a safe value ($\alpha_{safe}$) as described in Equation \ref{safety term}. Finally, we constrain the vehicle to remain within the lane boundaries, this gives us an admissible state space $X_l$. This has been implemented by spline fitting on the local lane boundaries as observed from the camera or from the pre-mapped region if the localization is accurate and setting the constraint that the position ($x_t, y_t$) falls between the left and right boundary splines. The resultant optimization objective is described as below in Equation \ref{total cost term}, where $\Delta t$ is the time step, and $\theta_{start}$ is the start state which is passed as the current state of the vehicle, $N$ is the no of time steps, and $F$ is the vehicle model function explained in Section \ref{Vehicle model}, $k_{pt}$, $k_{ot}$, $k_{dt}$, $k_{st}$ are the constants to scale the individual cost terms described above.
\begin{flalign}
    \begin{split}\label{safety term}
    C_{st} &= (\alpha_F > \alpha_{safe}) (\alpha_F  - \alpha_{safe})^2 + \\ 
    &\quad\quad\quad\quad\quad(\alpha_R > \alpha_{safe}) (\alpha_R - \alpha_{safe})^2
    \end{split}
\end{flalign}
\begin{flalign}
    \begin{split}\label{total cost term}
        \mathop{\min}\limits_{\theta , u}&\quad\sum_{t=0}^{N-1} k_{pt} C_{pt} + k_{ot} C_{ot} + k_{dt} C_{dt} + k_{st} C_{st}\\
        s.t.&\quad\theta_{0} - \theta_{start} = 0,\\
        &\quad \theta_{t+1} - \Delta t F(\theta_{t},u_t) = \theta_t\\
        &\quad \theta_t \in X_l
    \end{split}
\end{flalign}
\subsection{Algorithm}

The final algorithm is briefly described in Algorithm \ref{Final control algorithm}. We maintain a list of active obstacles to be passed onto NMPC and also maintain the state for each of the dynamic obstacles where we store the previously decided direction for overtaking. This will avoid possible anomaly due to numeric errors when the conditions are symmetric, and the cost for overtaking from right or left is similar for the target vehicle. We take the $N-1$ planned steps from the previous NMPC step and check if, for them, the ego vehicle comes in the vicinity of any opponent vehicle. If so, the overtaking strategy described above (Section \ref{Overtaking Strategy}) is used to decide the direction of overtaking, and the initial trajectory is set up accordingly to be passed to NMPC for optimization. The first step on the resultant trajectory is executed, and the process is repeated.\\

\alglanguage{pseudocode}
\begin{algorithm}
\small
\caption{Real-time NMPC algorithm}
\label{Final control algorithm}
\begin{algorithmic}[1]
    \State state\_opp $\gets []$
    \While{\textbf{TRUE}}
    \State opp\_detected$\gets$ all detectable opponents
    \State get current position and velocities
    \Function{MpcCallBack}{}
        \parState{%
        $N$-1  planned  steps  from  the  previous  NMPC  step  are checked.}
        \If {ego  vehicle  comes  in  vicinity to any opponent vehicle \hspace*{5.7em}in opp\_detected during any step.}
            \State target\_index $\gets$ index of closest vehicle
            \State target\_id $\gets$ opp\_detected[target\_index]
            \State pref\_direction $\gets$ null
            \If{closest opponent same as closest opponent in \hspace*{7.3em}previous iteration}
                \State pref\_direction $\gets$ state\_opp[target\_id] 
            \Else
                \State state\_opp $\gets []$
            \EndIf
            \State traj\_init $\gets \text{overtaking strategy}$
            \State state\_opp[target\_id] $\gets \text{Overtake Direction}$
            \Statex \Comment{Overtake Direction = right or left}
        \Else 
            \State traj\_init $\gets \text{Straight Trajectory}$
            \State state\_opp $\gets \text{[]}$
        \EndIf
        \State traj\_init is optimised using IPOPT
    \EndFunction
    \State \text{send first control at the end of time slot.}
    \EndWhile
\Statex
\end{algorithmic}
  \vspace{-0.4cm}%
\end{algorithm}
\section{Results} \label{Results}
\begin{figure}[h!]
        \centering\includegraphics[width=4.5cm]{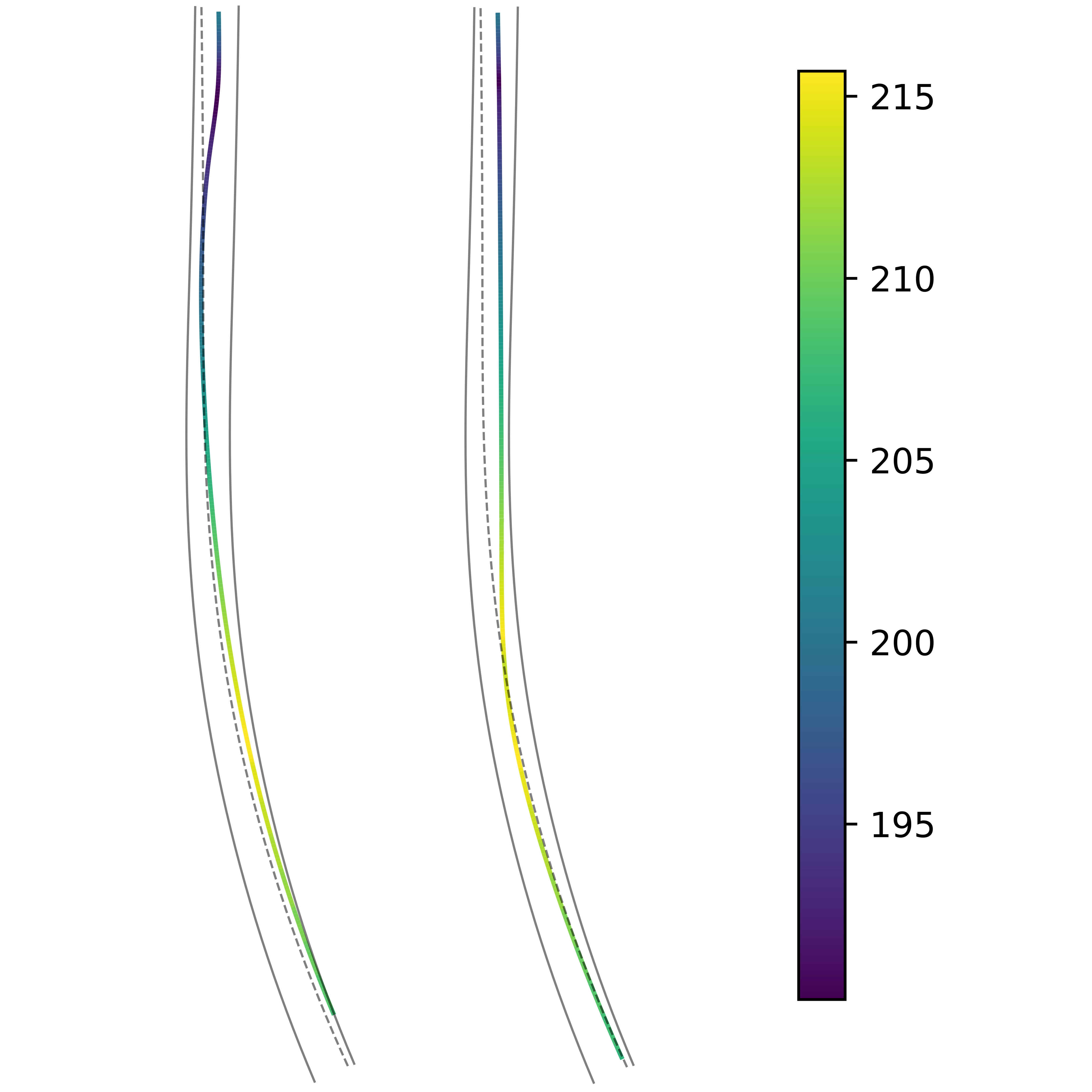}
      \caption{Dotted trajectory is the optimal racing line, hard lines are the lane boundaries Left : Local trajectory followed with contouring error formulation, Right: Local trajectory followed with new cost term. The color-bar represents the speeds in km/hr}
      \label{results 1}
  \end{figure}

    In order to support the argument that is using the new cost term for making more progress along the track using the optimal racing line, we compare the qualitative results using the new term with the contouring error term as used in \cite{Liniger_2014} in Fig. \ref{results 1}. From the results, it can be empirically put that with the new cost term, as the merging happens at the time of turns, the net distance covered is less, and also, the term does not interfere with the overtaking objective during straight regions of the track.
    \begin{figure}[h!]
    \centering\includegraphics[width=4.5cm]{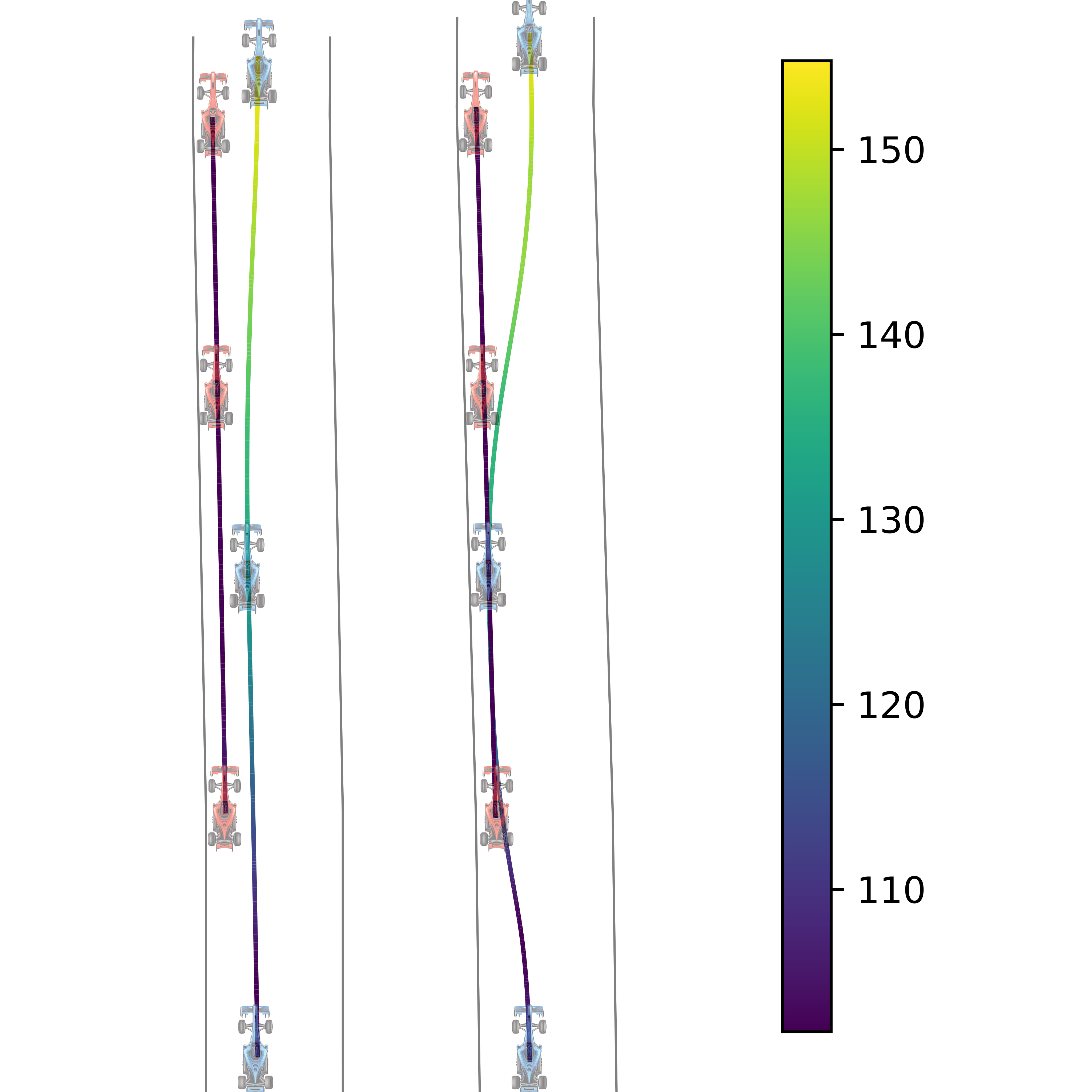}
      \caption{Left : Local trajectory followed without drafting cost term, Right: Local trajectory followed with drafting cost term. The color-bar represents the speeds in km/hr}
      \label{results 2}
  \end{figure}
Secondly, to demonstrate the effect of using drafting cost term in the presence of drafting, we compare the results with and without using drafting cost term as demonstrated in Fig. \ref{results 2} using the coefficient values: $k_c$ = 0.805, $k_x$ = 0.003, $k_y$ = 0.0825 (refer to Section \ref{Drafting model}). Finally, we attach a link to a video \footnote[3]{\href{https://youtu.be/PhScm6MnrRM}{https://youtu.be/PhScm6MnrRM} } demonstrating a 5 lap run using the controller on a multi agent racing scenario which was recorded during the first practice round as part of IAC. However, the simulation is run synchronously such that the algorithm works at 25 FPS in simulation time. The simulation time is slower than real time by an average factor of 5. Hence, the algorithm works with a mean frame rate of 5 FPS in real time on Intel(R) Xeon(R) CPU E5-2690 v3@2.60GHz 
. 
\section{Conclusion}

In conclusion, the algorithm proposed can compete considering multiple objectives while racing and can use the optimal racing line for making decisions effectively. It uses numeric optimization on an initial decision path to consider multiple objectives like collision avoidance and effectively leverage the use of drafting for overtaking by staying behind the lead vehicle to gain more speed. The algorithm was tested on ANSYS VRXPERIENCE simulator, and it satisfactorily follows the objectives of the multi-agent head-to-head racing scenarios. However, the algorithm runs at 5 FPS in real-time with horizon length as 1s which is quite slow for real time deployment. Optimization takes longer using IPOPT as expected due to the non linear nature of the model. The computation time can be reduced considerably by linearizing the model at each time step to form an LTV system. This can be solved efficiently using dynamic programming based solutions like iterative LQR. This would drastically reduce the computation time for real time deployment which we leave as future work.

\section*{Acknowledgment}
This work was done as part of the Indy Autonomous Challenge (IAC) under the collaborative IUPUI-IITKGP-USB team. Special thanks to all the team members, Adarsh Patnaik and Qin Lin for their support and guidance.





\printbibliography

\end{document}